\documentclass{article}

\usepackage[nonatbib,final]{neurips_2020}
\usepackage[utf8]{inputenc}
\usepackage[T1]{fontenc}
\usepackage[pagebackref=true,colorlinks]{hyperref}
\usepackage{url}
\usepackage{booktabs}
\usepackage{amsfonts}
\usepackage{nicefrac}
\usepackage{microtype}

\usepackage{multirow}
\usepackage[font={small}]{caption}
\usepackage[labelformat=simple]{subcaption}

\usepackage{tablefootnote}
\usepackage[dvipsnames]{xcolor}
\usepackage{graphicx}
\usepackage{pdfpages}
\usepackage{enumitem}

\title{Self-Supervised Learning by Cross-Modal Audio-Video Clustering}

\author{
Humam Alwassel$^{1\thanks{Work done during an internship at Facebook AI}}$ \\
\texttt{\small humam.alwassel@kaust.edu.sa} \\ \And
Dhruv Mahajan$^{2}$ \\
\texttt{\small dhruvm@fb.com} \\ \And
Bruno Korbar$^{2}$ \\
\texttt{\small bkorbar@fb.com} \\ \AND
Lorenzo Torresani$^{2}$ \\
\texttt{\small torresani@fb.com} \\ \And
Bernard Ghanem$^{1}$ \\
\texttt{\small bernard.ghanem@kaust.edu.sa} \\ \And
Du Tran$^{2}$ \\
\texttt{\small trandu@fb.com} \\ 
\AND \vspace{-16pt}
\\
$^{1}$King Abdullah University of Science and Technology (KAUST) \quad $^{2}$Facebook AI \\
{\url{http://humamalwassel.com/publication/xdc}}
}

\begin{document}

\maketitle

\begin{abstract}
Visual and audio modalities are highly correlated, yet they contain different information. Their strong correlation makes it possible to predict the semantics of one from the other with good accuracy. Their intrinsic differences make cross-modal prediction a potentially more rewarding pretext task for self-supervised learning of video and audio representations compared to within-modality learning. Based on this intuition, we propose \emph{Cross-Modal Deep Clustering (XDC)}, a novel self-supervised method that leverages unsupervised clustering in one modality (\emph{e.g.}, audio) as a supervisory signal for the other modality (\emph{e.g.}, video). This cross-modal supervision helps XDC utilize the semantic correlation and the differences between the two modalities. Our experiments show that XDC outperforms single-modality clustering and other multi-modal variants. XDC achieves state-of-the-art accuracy among self-supervised methods on multiple video and audio benchmarks. Most importantly, our video model pretrained on large-scale unlabeled data significantly outperforms the same model pretrained with full-supervision on ImageNet and Kinetics for action recognition on HMDB51 and UCF101. To the best of our knowledge, XDC is the first self-supervised learning method that outperforms large-scale fully-supervised pretraining for action recognition on the same architecture.
\end{abstract}
\section{Introduction}\label{sec:intro}

Do we need to explicitly name the actions of ``laughing'' or ``sneezing'' in order to recognize them? Or can we learn to visually classify them without labels by associating characteristic sounds with these actions? Indeed, a wide literature in perceptual studies provides evidence that we rely heavily on hearing sounds to make sense of actions and dynamic events in the visual world. For example, objects moving together are perceived as bouncing off each other when the visual stimulus is accompanied by a brief sound~\cite{crossmodal}, and the location and timing of sounds are leveraged as important cues to direct our spatiotemporal visual attention~\cite{Heffner1992,Naatanen1992}. The influence of hearing sounds in visual perception is also suggested by perceptual studies showing that individuals affected by profound deafness exhibit poorer visual perceptual performance compared to age-matched hearing controls~\cite{gentile1969academic,myklebust1960psychology}. 

In this work, we investigate the hypothesis that spatiotemporal models for action recognition can be reliably pretrained from {\em unlabeled} videos by capturing cross-modal information from audio and video. The motivation for our study stems from two fundamental challenges facing a fully-supervised line of attack to learning video models. The first challenge is the exorbitant cost of scaling up the size of manually-labeled video datasets. The recent creation of large-scale action recognition datasets~\cite{caba2015activitynet,something_something,Sports1M,kinetics} has undoubtedly enabled a major leap forward in video models accuracies. However, it may be argued that additional significant gains by dataset growth would require scaling up existing labeled datasets by several orders of magnitude. 
The second challenge is posed by the unclear definition of suitable label spaces for action recognition. Recent video datasets differ substantially in their label spaces, which range from sports actions~\cite{Sports1M} to verb-noun pairs for kitchen activities~\cite{Damen2018EPICKITCHENS}. This suggests that the definition of the ``right'' label space for action recognition, and more generally for video understanding, is still very much up for debate. It also implies that finetuning models pretrained on large-scale labeled datasets is a suboptimal proxy for learning models for small- or medium-size datasets due to the label-space gap often encountered between source and target datasets. 

\vspace{-2pt}
In this paper, we present three approaches for training video models from self-supervised audio-visual information. At a high-level, the idea behind all three frameworks is to leverage one modality (say, audio) as a supervisory signal for the other (say, video). We posit that this is a promising avenue because of the simultaneous synergy and complementarity of audio and video: correlations between these two modalities make it possible to perform prediction from one to the other, while their intrinsic differences make cross-modal prediction an enriching self-supervised task compared to within-modality learning. Specifically, we adapt the single-modality DeepCluster work of Caron~\emph{et al.}~\cite{caron2018deep} to our multi-modal setting. DeepCluster was introduced as a self-supervised procedure for learning image representation. It alternates between unsupervised clustering of image features and using these cluster assignments as pseudo-labels to revise the image representation. 
In our work, the clusters learned from one modality are used as pseudo-labels to refine the representation for the other modality. In two of our approaches---Multi-Head Deep Clustering (MDC) and Concatenation Deep Clustering (CDC)---the pseudo-labels from the second modality are {\em supplementary}, \emph{i.e.}, they complement the pseudo-labels generated in the first modality. The third approach---Cross-Modal Deep Clustering (XDC)---instead uses the pseudo-labels from the other modality as an {\em exclusive} supervisory signal. This means that in XDC, the audio clusters drive the learning of the video representation and vice versa. 
Our experiments support several interesting conclusions:
\vspace{-5pt}
\begin{itemize}[nosep, leftmargin=.42cm]
\item All three of our cross-modal methods yield representations that generalize better to the downstream tasks of action recognition and audio classification, compared to their within-modality counterparts. 
\item XDC (\emph{i.e.}, the cross-modal deep clustering relying on the other modality as an exclusive supervisory signal) outperforms all the other approaches. This underscores the complementarity of audio and video and the benefits of learning label-spaces across modalities.
\item Self-supervised cross-modal learning with XDC on a large-scale video dataset yields an action recognition model that achieves higher accuracy when finetuned on HMDB51 or UCF101, compared to that produced by fully-supervised pretraining on Kinetics. To the best of our knowledge, this is the first method to demonstrate that self-supervised video representation learning outperforms large-scale fully-supervised pretraining for action recognition. 
Moreover, unlike previous self-supervised methods that are only pretrained on curated data (\emph{e.g.}, Kinetics~\cite{kinetics} without action labels), we also report results of XDC pretrained on a large-scale uncurated video dataset.
\end{itemize}

\vspace{-5pt}
\section{Related work}
\vspace{-2pt}
\label{sec:related_work}

\vspace{-3pt}\noindent{\bf Early unsupervised representation learning.}
Pioneering works include deep belief networks~\cite{hinton2006fast}, autoencoders~\cite{hinton2006reducing,vincent2008extracting}, shift-invariant decoders~\cite{ranzato2007unsupervised}, sparse coding algorithms~\cite{lee2007efficient}, and stacked ISAs~\cite{le2011learning}. While these approaches learn by reconstructing the input, our approach learns from a self-supervised pretext task by generating pseudo-labels for supervised learning from unlabeled data.

\vspace{-3pt}\noindent{\bf Self-supervised representation learning from images and videos.}
Several pretext tasks exploit image spatial context, \emph{e.g.}, by predicting the relative position of patches~\cite{doersch2015unsupervised} or solving jigsaw puzzles~\cite{noroozi2016unsupervised}. 
Others include creating image classification pseudo-labels (\emph{e.g.}, through artificial rotations~\cite{gidaris2018unsupervised} or clustering features~\cite{caron2018deep}), colorization~\cite{zhang2016colorful}, inpainting~\cite{pathak2016context}, motion segmentation~\cite{pathak2017learning}, and instance counting~\cite{noroozi2017representation}.
Some works have extended image pretext tasks to video~\cite{kim2019self,motion_statistics,clip_order}. Other video pretext tasks include frame ordering~\cite{fernando2017self,lee2017unsupervised,misra2016shuffle,wei2018learning}, predicting flow or colors~\cite{lai2019self,vondrick2018tracking}, exploiting region correspondences across frames~\cite{isola2015learning,jayaraman2016slow,wang2015unsupervised,wang2019learning}, future frame prediction \cite{lotter2016deep,mathieu2015deep,srivastava2015unsupervised,vondrick2016anticipating,vondrick2016generating}, and tracking~\cite{XiaolongWang19}. Unlike this prior work, our model uses two modalities: video and audio.

\vspace{-3pt}\noindent{\bf Cross-modal learning and distillation.} 
Several works~\cite{aytar2016soundnet,gupta2016cross} train a fully-supervised encoder on one modality and distill its discriminative knowledge to an encoder of a different modality.
Other works learn from unlabeled data for a specific target task~\cite{zhao2018sound,rouditchenko2019self}. 
Unlike these methods, our work is purely self-supervised and aims at learning representations that transfer well to a wide range of downstream tasks.
Previous cross-modal self-supervised methods most relevant to our work include audio-visual correspondence~\cite{Arandjelovic17}, deep aligned representations~\cite{aytar2017see}, audio-visual temporal synchronization~\cite{avts,owens2018audio}, contrastive multiview coding~\cite{tian2020contrastive}, and learning image representations using ambient sound~\cite{owens2016ambient}.
While~\cite{Arandjelovic17,aytar2017see,owens2016ambient,tian2020contrastive} use only a single frame, we use a video clip. Unlike our method, \cite{owens2016ambient} clusters handcrafted audio features and does not iterate on the pseudo-labels.
\cite{avts,owens2018audio} require constructing positive/negative examples for in-sync and out-of-sync video-audio pairs. This sampling strategy makes these approaches more difficult to scale compared to ours, as many potential out-of-sync pairs can be generated, yielding largely different results depending on the sampling choice~\cite{avts}. 
Recent works, such as MIL-NCE~\cite{miech2020end} and CBT~\cite{sun2019learning}, learn from unlabeled instructional videos using text from ASR, while our approach makes use of the audio signal instead.

\begin{figure}[t]
  \centering
  \includegraphics[width=\linewidth]{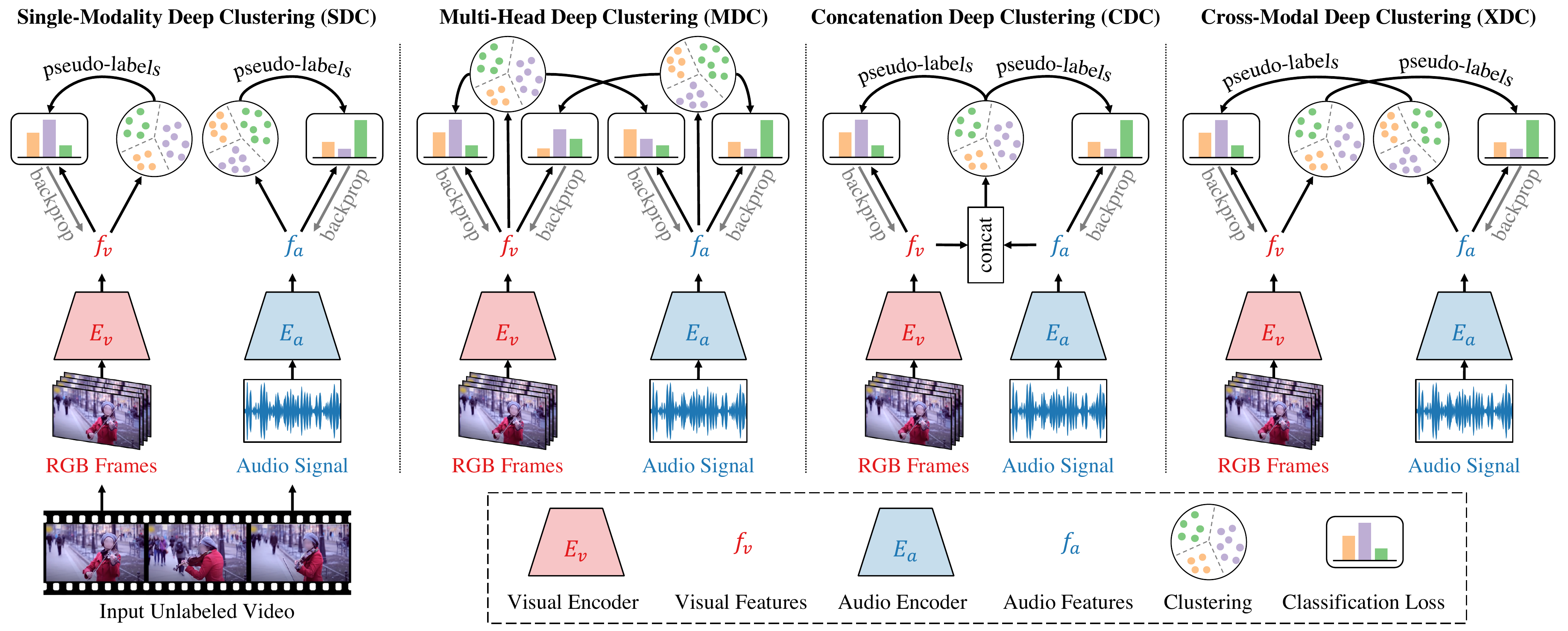}
  \vspace{-16pt}
  \caption{\textbf{Overview of our framework.} We present Single-Modality Deep Clustering (SDC) baseline vs. our three multi-modal deep clustering models: Multi-Head Deep Clustering (MDC), Concatenation Deep Clustering (CDC), and Cross-Modal Deep Clustering (XDC). 
  The video and audio encoders ($E_v$ and $E_a$) map unlabeled videos to visual and audio features ($f_v$ and $f_a$). These features, or their concatenations, are clustered using $k$-means. The cluster assignments are then used as pseudo-labels to train the encoders. We start with randomly-initialized encoders, then alternates between clustering to generate pseudo-labels and training to improve the encoders. The four models employ different ways to cluster features and generate self-supervision signals. Illustration video is from ~\cite{cc_video_pictures}.} 
  
  \vspace{-8pt}
  \label{fig:pipeline_fig}
\end{figure}

\vspace{-1pt}
\section{Technical approach}
\vspace{-3pt}

\label{sec:technical_appraoch}
Here, we briefly discuss previous work on single-modality deep clustering in images~\cite{caron2018deep}. Then, we introduce our three multi-modal deep clustering frameworks for representation learning (Figure~\ref{fig:pipeline_fig}).

\vspace{-3pt}
\subsection{Single-modality deep clustering}
\vspace{-3pt}

Caron~\emph{et al.}~\cite{caron2018deep} proposed DeepCluster for self-supervised representation learning from images. DeepCluster iteratively clusters deep features from a single-modality encoder, and then uses the cluster assignments to train the same encoder to improve its representation. Inspired by the simplicity of this work, our paper studies deep clustering in the large-scale multi-modal setting.  
For completeness, we summarize DeepCluster details.
Let $X$ be the set of unlabeled inputs (\emph{e.g.}, images), $E$ be an encoder that maps an input $x \in X$ to a deep feature vector $f \in \mathbb{R}^d$. DeepCluster iterates between clustering the features $F = \{f = E(x) ~|~ x \in X\}$ and discriminative training to improve $E$ using the clustering assignments as pseudo-labels. The process starts with a randomly-initialized $E$, and only the weights of the classification \texttt{fc}-layer are reset between clustering iterations when the supervision-taxonomy is switched. DeepCluster uses a 2D CNN (\emph{e.g.} ResNet-$50$) for $E$ and clusters the features after each epoch using $k$-means. We refer to DeepCluster as \textbf{Single-Modality Deep Clustering (SDC)}.

\vspace{-2pt}
\subsection{Multi-modal deep clustering}
\vspace{-3pt}

Contrary to the single-modality case, there exist multiple encoders in a multi-modal setting, each of which encodes a different modality of the input. In our paper, we consider two modalities, the visual and the audio modalities from the unlabeled training video clips. In particular, let $X$ be the set of unlabeled video clips, and $E_v$ and $E_a$ be the visual and audio encoders, respectively.  Let $F_v = \{f_v = E_v(x) \in \mathbb{R}^{d_v} ~|~ x \in X\}$ and $F_a = \{f_a = E_a(x) \in \mathbb{R}^{d_a} ~|~ x \in X\}$ be the set of visual and audio deep features produced by the two encoders, respectively. There are different ways we can adapt the deep clustering framework to a multi-modal input. We describe three approaches (MDC, CDC, and XDC) by detailing the steps taken at each deep clustering iteration. Refer to the \emph{supplementary material} for the implementation differences between SDC and our three approaches.

\noindent\textbf{Multi-Head Deep Clustering (MDC).} 
This model builds on SDC by adding a second classification head supervised by the other modality. Thus, in this model, each encoder has two classification heads. At each deep clustering iteration, MDC uses the cluster assignments of $F_v$ as pseudo-labels for one head and that of $F_a$ as pseudo-labels for the other head. Thus, each encoder needs to predict the cluster assignments of its own modality (as in SDC), but also those generated by the other modality.

\noindent\textbf{Concatenation Deep Clustering (CDC).} 
This model performs clustering of joint visual and audio features. Specifically, at each deep clustering iteration, CDC clusters vectors obtained by concatenating the visual and audio feature vectors, separately $l_2$-normalized. Then, it uses the resulting cluster assignments as pseudo-labels to update the weights of both $E_v$ and $E_a$.

\noindent\textbf{Cross-Modal Deep Clustering (XDC).} 
Each encoder in this model relies exclusively on the clusters learned from the other modality as the supervisory signal. At each deep clustering iteration, XDC clusters the audio deep features, $F_a$, and uses their cluster assignments as pseudo-labels to train the visual encoder, $E_v$. Vice versa, XDC supervises $E_a$ with the cluster assignments of $F_v$.

\vspace{-2pt}
\section{Experiments}
\label{sec:ablation_experiments}

\vspace{-2pt}
\subsection{Experimental setup}
\vspace{-2pt}

\noindent {\bf Pretraining datasets.}
We use four datasets: Kinetics~\cite{kinetics}, AudioSet~\cite{AudioSet}, IG-Kinetics~\cite{ghadiyaram2019large}, and IG-Random, which have $240$K, $2$M, $65$M, and $65$M training videos, respectively.
As our approach is self-supervised, thus the labels from the first three datasets are {\bf not used} during pretraining. While Kinetics and AudioSet are supervised benchmarks for action recognition and audio classification, IG-Kinetics is a weakly-supervised dataset collected from a social media website using tags related to Kinetics actions. IG-Random is an \textit{uncurated} dataset of random videos from the same website. Videos are $10$-second long in Kinetics and AudioSet and $10$-to-$60$-second long in IG-Kinetics and IG-Random. We filter out around $7$K Kinetics videos that have no audio. Furthermore, we randomly sample $240$K videos from AudioSet and denote this subset as AudioSet-$240$K. We generate this subset to have AudioSet data of the same size as Kinetics, in order to study the effects of pretraining with the same data size but on a different data distribution and domain. 

\noindent {\bf Downstream datasets.}
We evaluate our pretraining performance on three downstream benchmarks: UCF101~\cite{UCF101}, HMBD51~\cite{HMDB51}, and ESC50~\cite{esc50}, which have $13$K, $7$K, and $2$K examples from $101$, $51$, and $50$ classes, respectively. UCF101 and HMDB51 are action recognition datasets, while ESC50 is a sound classification dataset. UCF101 and HMDB51 have $3$ official train/test splits, while ESC50 has $5$ splits. We conduct our ablation study (Subsection~\ref{sec:ablation_study}) using split-1 of each dataset. We also report our average performance over all splits when we compare with state-of-the-art methods in Section~\ref{sec:compare_sota}.

\noindent {\bf Baselines.}
We consider two baselines: \textit{{Scratch}} and \textit{{Supervised Pretraining (Superv)}}. The first is a randomly-initialized model trained from scratch directly on the downstream task, while the second is a model pretrained in a supervised fashion on a large labeled dataset (\emph{e.g.}, Kinetics) and then finetuned on the downstream task. We note that these two baselines are commonly regarded as the lower and upper bounds to gauge the quality of self-supervised representation learning methods~\cite{Arandjelovic17,avts}.

\noindent {\bf Backbone architectures.}
We employ R(2+1)D~\cite{Tran18} and ResNet~\cite{KaimingHe16} as $E_v$ and $E_a$, respectively. $E_v$'s input is a $3$$\times$$L$$\times$$H$$\times$$W$ clip, where $3$ refers to the RGB channels, $L$ is the number of frames, and $H$ and $W$ are the frame height and width. $E_a$'s input is a $Q$$\times$$P$ spectrogram image extracted from the audio signal, where $Q$ is the number of MEL filters and $P$ is the number of audio frames. 

\noindent {\bf Pretraining and evaluation details.}
We choose the $18$-layer variants of R(2+1)D and ResNet encoders. We use clips of $L$=$8$ frames for pretraining and finetuning our visual encoder $E_v$. We scale frames such that the smallest dimension is $256$ pixels and then random crop images of size $224$$\times$$224$. We extract video clips at $30$ fps and employ temporal jittering during training. 
For the audio input, we sample $2$ seconds and use $Q$=$40$ MEL filters and $P$=$100$ audio frames. For inference on the downstream tasks, we uniformly sample $10$ clips per testing example and average their predictions to make a video-level prediction. We use only one crop per clip: the center $8$$\times$$224$$\times$$224$ crop for video and the full $40$$\times$$100$ crop for audio. We provide more details in the \emph{supplementary material}.

\vspace{-1pt}
\subsection{Ablation study}\label{sec:ablation_study}
\vspace{-1pt}

\begin{table}[t!]
    \small
    \centering
    \caption{\textbf{Single-modality vs. multi-modal deep clustering.} We compare the four self-supervised deep clustering models (Section \ref{sec:technical_appraoch}) and the three baselines: Scratch, Supervised Pretraining (Superv), and same-modality-XDC (XDC with two encoders defined on the same modality). Models are pretrained via self-supervision on Kinetics and fully finetuned on each downstream dataset. We report the top-1 accuracy on split-1 of each dataset. All multi-modal models significantly outperform the single-modality deep clustering model. We mark in bold the best and underline the second-best models.}
    \begin{tabular}{l | c c | c c c c | c c }
        \hline
                &         &        &     &      &                  &           & \multicolumn{2}{c}{same-modality-XDC} \\
        Dataset & Scratch & Superv & SDC & MDC  & CDC              & XDC       & 2 visual encoders & 2 audio encoders \\
        \hline 
        UCF101 & 54.5    & 90.9   & 61.8 & 68.4 & \underline{72.9} & \bf{74.2} & 61.3 & N/A  \\
        HMDB51 & 24.1    & 58.0   & 31.4 & 37.1 & \underline{37.5} & \bf{39.0} & 30.5 & N/A \\
        ESC50  & 54.3    & 82.3   & 66.5 & 70.3 & \underline{74.8} & \bf{78.0} & N/A  & 66.0  \\
        \hline 
    \end{tabular}
    \vspace{-8pt}
    \label{table:ablation_study_dc_models}
\end{table}

\noindent {\bf Study 1: Single-modality vs. multi-modal deep clustering.}
This experiment compares the four models presented in Section~\ref{sec:technical_appraoch}. We pretrain SDC, MDC, CDC, and XDC on Kinetics and report their performance on the downstream tasks in Table~\ref{table:ablation_study_dc_models}. 
To better understand XDC, we also conduct a new set of baselines, called same-modality-XDC, where XDC is trained with two encoders defined on the \textit{same} modality (either visual or audio).
Note that all models in this ablation study use the same visual and audio encoders and only differ in the way they use self-supervision. It takes on average $5$ to $6$ deep clustering iterations for these models to converge. 
\textbf{\textit{Observations:}}
\textbf{{(I)}} The four self-supervised deep clustering models outperform the Scratch baseline on all downstream benchmarks. This shows that our self-supervised pretraining is effective and generalizes well to multiple tasks.
\textbf{{(II)}} All multi-modal models (MDC, CDC, and XDC) significantly outperform SDC by up to $12.4$\%, $7.6$\%, and $11.5$\% on UCF101, HMDB51, and ESC50, respectively. This validates the importance of multi-modal modeling compared to single-modality.
\textbf{{(III)}} XDC achieves the best performance across all tasks. What distinguishes XDC from the other models is that each modality encoder in XDC is self-supervised purely by the signal from the other modality. The encoders in CDC, MDC, and SDC all employ a self-supervision signal coming from the same modality. Thus, this suggests that encoders learn better when purely supervised by a different modality. 
We provide the following intuition on why XDC is better than CDC and MDC. XDC groups samples together when they are similar in one of the two modalities (video to supervise the audio encoder, audio to supervise the visual encoder). Instead, CDC groups samples together only if they are similar according to both the audio \textit{and} the video modality (to supervise both encoders). Thus, XDC visual and audio clusters allow for more diversity than those of CDC. We hypothesize that this diversity allows XDC to learn richer representations, which translates into better performance on the downstream tasks. Also, recent work~\cite{Wang_2020_CVPR} has shown that models trained on different modalities learn and generalize at different speeds, and that training them jointly (as done in MDC which uses two-modality heads) is sub-optimal. We believe that this could contribute to MDC performing worse than XDC, which optimizes for each modality independently.
\textbf{{(IV)}} The same-modality-XDC baselines perform similarly to SDC and are $8$-$12$\% worse than multi-modal-XDC. This suggests that cross-modality provides a superior supervisory signal for self-supervised learning and that multi-modal-XDC is the best model not because of its optimization strategy but rather because of the use of the other modality for pseudo-labeling. Given the results of this study, we opt to use only XDC in the rest of the experiments.
Finally, to show that XDC works for different backbones, we re-do Study 1 with ResNet3D in the \emph{supplementary material}.

\noindent {\bf Study 2: The number of $k$-means clusters.} 
This study explores the effects of changing the hyperparameter $k$ in $k$-means clustering. We pretrain XDC on three datasets, Kinetics, AudioSet-$240$K, and AudioSet, using $k$=$64, 128, 256, 512,$ and $1024$ clusters (Table~\ref{table:ablation_study_k}). 
\textbf{\textit{Observations:}}
\textbf{{(I)}} The best $k$ value is not sensitive to the number of semantic labels in the downstream datasets. For example, HMDB51 and ESC50 have about the same number of labels but different best $k$ value.
\textbf{{(II)}} Similarly, the best $k$ value seems uncorrelated with the number of original semantic labels of the pretraining dataset, \emph{e.g.} $400$ in Kinetics. We reiterate here that our approach is self-supervised and \textbf{does not use} the labels of the pretraining dataset.
\textbf{{(III)}} The best $k$ value tends to get larger as the pretraining data size increases. For example, the best $k$ for HMDB51 shifts from $128$ to $256$ when moving from pretraining on AudioSet-$240$K to the full AudioSet. We hypothesize that there is a more diverse sample set to cluster when the pretraining data size increases. Thus, we can have more fine-grained clusters (higher $k$) and make our self-supervised classification problem harder. This aligns with previous self-supervised works~\cite{goyal2019scaling,avts} that showed benefits from making the pretext task harder.
\begin{table}[t!]
    \small
    \centering
    \caption{\textbf{The number of clusters ($k$).} We show the effect of the number of $k$-means clusters on XDC performance. XDC is pretrained on three large datasets, and then fully finetuned on three downstream tasks. We report the top-1 accuracy on split-1. The best $k$ value increases as the size of the pretraining dataset increases.}
    \begin{tabular}{l | l | c c c c c }
        \hline
        \multirow{2}{7em}{Pretraining\\Dataset} & \multirow{2}{6em}{Downstream\\Dataset} & \multicolumn{5}{c}{$k$} \\
        & & 64 & 128 & 256 & 512 & 1024 \\ 
        \hline
        \multirow{3}{7em}{Kinetics\\($240$K videos)} & UCF101 & {73.8} & 73.1 & \bf{74.2} & \underline{74.0} & 72.6 \\
        & HMDB51 & 36.5 & \bf{39.0} & \underline{38.3} & 37.7 & 37.7 \\
        & ESC50  & \bf{78.0} & \underline{76.3} & 75.0 & 74.5 & 71.5 \\
        \hline
        \multirow{3}{7em}{AudioSet-240K\\($240$K  videos)} & UCF101 & {\bf 77.4} & \underline{77.2} & 76.7 & 77.1 & 75.3 \\
        & HMDB51 & 41.3 & {\bf 42.6} & \underline{41.6} & 40.6 & 40.7 \\
        & ESC50  & {\bf 78.5} & \underline{77.8} & 77.3 & 76.8 & 73.5 \\
        \hline
        \multirow{3}{7em}{AudioSet\\($2$M videos)} & UCF101 & 84.1 & 84.3 & \bf{84.9} & \underline{84.4} & 84.2 \\
        & HMDB51 & 47.4 & 47.6 & \bf{48.8} & \underline{48.5} & 48.4 \\
        & ESC50  & 84.8 & \bf{85.8} & \underline{85.0} & 84.5 & 83.0 \\
        \hline
    \end{tabular}
    \vspace{-8pt}
    \label{table:ablation_study_k}
\end{table}

\begin{table}[t!]
    \small
    \centering
    \caption{\textbf{Pretraining data type and size.} We compare XDC pretrained on five datasets vs. fully-supervised pretrained baselines (Superv). XDC significantly outperforms fully-supervised pretraining on HMDB51.}
    \begin{tabular}{l l l | c c c }
        \hline
        \multicolumn{3}{c|}{Pretraining} & \multicolumn{3}{c}{Downstream Dataset} \\
        Method & Dataset & Size & UCF101 & HMDB51 & ESC50 \\
        \hline 
        Scratch &  None & 0 & 54.5 & 24.1 & 54.3\\
        \hline
        Superv & ImageNet & 1.2M & 79.9 & 44.5 & NA \\
        Superv & Kinetics & 240K & \underline{90.9} & 58.0 & 82.3\\
        Superv & AudioSet-240K & 240K  & 76.6 & 40.8 & 78.3\\
        Superv & AudioSet & 2M & 84.0 & 53.5 & {\bf 90.3}\\
        \hline
        XDC & Kinetics & 240K & 74.2 & 39.0 & 78.0\\
        XDC & AudioSet-240K & 240K & 77.4 & 42.6 & 78.5\\
        XDC & AudioSet & 2M & 84.9 & 48.8 & 85.8\\
        XDC & IG-Random   & 65M & 88.8 & \underline{61.2} & \underline{86.3}\\
        XDC & IG-Kinetics & 65M & {\bf 91.5} & {\bf 63.1} & 84.8 \\
        \hline
    \end{tabular}
    \vspace{-8pt}
    \label{table:ablation_study_pretrain_dataset_type_and_size}
\end{table}

\noindent {\bf Study 3: Pretraining data type and size.} 
Here, we investigate the effects of two pretraining characteristics: data size and type. To this end, we pretrain XDC on Kinetics ($240$K examples), AudioSet-$240$K ($240$K examples), AudioSet ($2$M examples), IG-Kinetics ($65$M examples), and IG-Random ($65$M examples). Kinetics and IG-Kinetics videos are collected originally for activity recognition, while AudioSet videos are aimed for audio event classification. IG-Random is an uncurated/unsupervised dataset. In addition to video datasets, we also experiment with ImageNet to understand how much action recognition benefits from supervised pretraining on object classification. For ImageNet, we inflate the images into static video clips (repeating the same frame) and pretrain our video model on this dataset. Table~\ref{table:ablation_study_pretrain_dataset_type_and_size} presents the results of this study.
\textbf{\textit{Observations:}} 
\textbf{{(I)}} XDC improves across all three downstream tasks as the pretraining data size increases. For example, XDC on HMDB51 improves by $9.8\%$, $22.2\%$, and $24.1\%$ when pretrained on AudioSet, IG-Random, and IG-Kinetics, respectively, compared to the results when pretrained on Kinetics.
\textbf{{(II)}} XDC outperforms Kinetics fully-supervised pretraining by $5.1\%$ on HMDB51 and by $0.6\%$ on UCF101. {To the best of our knowledge, XDC is the first method to demonstrate that self-supervision can outperform large-scale full-supervision in representation learning for action recognition}.
\textbf{{(III)}} The performance of the fully-supervised pretrained model is influenced by the taxonomy of the pretraining data more than the size. For example, supervised-pretraining on Kinetics gives better performance on both UCF101 and HMDB51 compared to supervised-pretraining on AudioSet (which is $8$ times larger than Kinetics) and ImageNet. One the other hand, XDC performance is less sensitive to the data type, as it implicitly learns the label space rather than depend on a space manually defined by annotators.

\noindent {\bf Study 4: Curated vs. uncurated pretraining data.} 
The overarching goal of self-supervised representation learning is to learn from the massive amounts of unlabeled data. Previous self-supervised methods have pretrained on videos from supervised (curated) datasets (\emph{e.g.}, Kinetics) without using the labels. However, even without using labels, those videos are still biased due to the sampling distribution (\emph{e.g.}, taxonomy of the curated dataset). To this end, we study the effects of self-supervised representation learning from uncurated data. Table~\ref{table:ablation_study_curated_vs_uncurated} compares XDC pretrained on IG-Kinetics (curated, as videos were tag-retrieved) vs. IG-Random (uncurated) using $1$M, $16$M, and $65$M videos.
\textbf{\textit{Observations:}} 
\textbf{{(I)}} Curated pretraining gives better results on UCF101 and HMDB51, while uncurated pretraining is better on ESC50 at large scale. We hypothesize that the bias of IG-Kinetics towards semantics of human actions is the reason behind the positive effect of curation on UCF101 and HMDB51. However, such bias negatively impacts the performance on ESC50. \textbf{{(II)}} The performance gap between the curated and uncurated pretraining shrinks significantly as we increase the data size. For example, the performance gap on HMDB51 drops from $5.2\%$ to $2.1\%$ and $1.9\%$ when the pretraining size increases from $1$M to $16$M and $65$M videos, respectively. This implies that XDC can learn meaningful representations from truly uncurated data. {To the best of our knowledge, XDC is the first self-supervised method to study pretraining on large-scale uncurated video data}.
\begin{table}[t!]
    \small
    \centering
    \tabcolsep=0.08cm
    \caption{\textbf{Curated vs. uncurated pretraining data.} XDC pretrained on IG-Kinetics (curated) vs. IG-Random (uncurated) using different training set sizes. Uncurated pretraining has better results on ESC at large scale. On UCF and HMDB, the accuracy gap between curated and uncurated pretraining decreases as data size increases.}
    \begin{tabular}{l | c c c | c c c | c c c }
        \hline
        \multirow{3}{5em}{Downstream\\Dataset} & \multicolumn{9}{c}{Pretraining Size} \\
        & \multicolumn{3}{c|}{\bf 1M} & \multicolumn{3}{c|}{\bf 16M} & \multicolumn{3}{c}{\bf 65M} \\
        & IG-Random & IG-Kinetics & $\Delta$ & IG-Random & IG-Kinetics & $\Delta$ & IG-Random & IG-Kinetics & $\Delta$ \\
        \hline
        UCF101 & 79.6 & {\bf 84.2} & -4.6 & 84.1 & {\bf 87.6} & -3.5 & 88.8 & {\bf 91.5} & -2.7 \\
        HMDB51 & 45.1 & {\bf 50.3} & -5.2 & 55.2 & {\bf 57.3} & -2.1 & 61.2 & {\bf 63.1} & -1.9 \\ 
        ESC50  & 77.8 & {\bf 79.5} & -1.7 & {\bf 84.3} & 82.5 & +1.8 & {\bf 86.3} & 84.8 & +1.5 \\
        \hline
    \end{tabular}
    \vspace{-8pt}
    \label{table:ablation_study_curated_vs_uncurated}
\end{table}

\begin{table}[t!]
    \small
    \centering
    \caption{{\bf Full finetuning vs. learning \texttt{fc}-only.} 
    We compare XDC against the supervised pretrained models (Superv) under two transfer-learning schemes: when models are used as features extractor (`\texttt{fc}' column) or as a finetuning initialization (`all' column). XDC fixed features outperform several fully-finetuned Superv models.}
    \begin{tabular}{l | l | c c | c c | c c }
        \hline
        \multirow{2}{4em}{Method} & \multirow{2}{5em}{Pretraining Dataset} & \multicolumn{2}{c}{UCF101} & \multicolumn{2}{|c}{HMDB51} & \multicolumn{2}{|c}{ESC50} \\
        & & \texttt{fc} & all & \texttt{fc} & all & \texttt{fc} & all \\ 
        \hline
        Random & None & {\scriptsize 6.0$\pm$1.0} & 54.5 & {\scriptsize 7.5$\pm$0.6} & 24.1 & {\scriptsize 61.3$\pm$2.5} & 54.3 \\
        \hline
        Superv & ImageNet & 74.5 & 79.9 & 42.8 & 44.5 & NA & NA \\
        Superv & Kinetics & {\bf 89.7} & \underline{90.9} & {\bf 61.5} & 58.0 & 79.5 & 82.3\\
        Superv & AudioSet & 80.2 & 84.0 & 51.6 & 53.5 & {\bf 88.5} & {\bf 90.3} \\
        \hline
        XDC & IG-Random & {80.7} & 88.8 & {49.9} & \underline{61.2} & {\underline{84.5}} & \underline{86.3}\\
        XDC & IG-Kinetics & \underline{85.3} & {\bf 91.5} & \underline{56.0} & {\bf 63.1} & 84.3 & 84.8\\
        \hline
    \end{tabular}
    \vspace{-8pt}
    \label{table:ablation_study_fc_only_vs_full_finetune}
\end{table}

\noindent {\bf Study 5: Full finetuning vs. learning \texttt{fc}-only.} 
Here, we study two approaches for transferring XDC to downstream tasks. 
\textit{Full finetuning}: we finetune all parameters of the pretrained encoder on the downstream task.
\textit{Learning \texttt{fc}-only}: we fix the pretrained encoder and learn a linear classifier for the downstream task, \emph{i.e.}, a fully-connected (\texttt{fc}) layer on top of the frozen features. Table~\ref{table:ablation_study_fc_only_vs_full_finetune} compares XDC with the supervised pretrained approaches under these two transfer-learning schemes.
\textbf{\textit{Observations:}}
\textbf{{(I)}} The accuracy of most pretrained models (fully-supervised or self-supervised) degrades, when used as a fixed feature extractor compared to when they are fully-finetuned on the downstream tasks. Nonetheless, the relative performance of XDC compared to supervised pretrained models stays generally the same when fully vs. \texttt{fc}-only finetuned on the downstream task. This suggests that XDC pretraining is useful both as a fixed feature extractor and as a pretraining initialization.
\textbf{{(II)}} XDC as a fixed feature extractor outperforms many fully-finetuned supervised models. For example, \texttt{fc}-only XDC outperforms, by significant margins, the fully-finetuned supervised AudioSet- and ImageNet-pretrained models on both UCF101 and HMDB51.
\textbf{{(III)}} We observe that fully-supervised pretraining, followed by \texttt{fc}-only finetuning performs well when the pretraining taxonomy is well aligned with that of the downstream task. For example, pretraining on Kinetics by learning \texttt{fc}-only on HMDB51 and UCF101 gives the best performance. This is expected as the label spaces of HMBD51 and UCF101 overlap largely with that of Kinetics. This suggests that fully-supervised pretraining is more taxonomy/downstream-task dependent, while our self-supervised XDC is taxonomy-independent.
\begin{figure}[t]
   \centering
   \includegraphics[width=\linewidth]{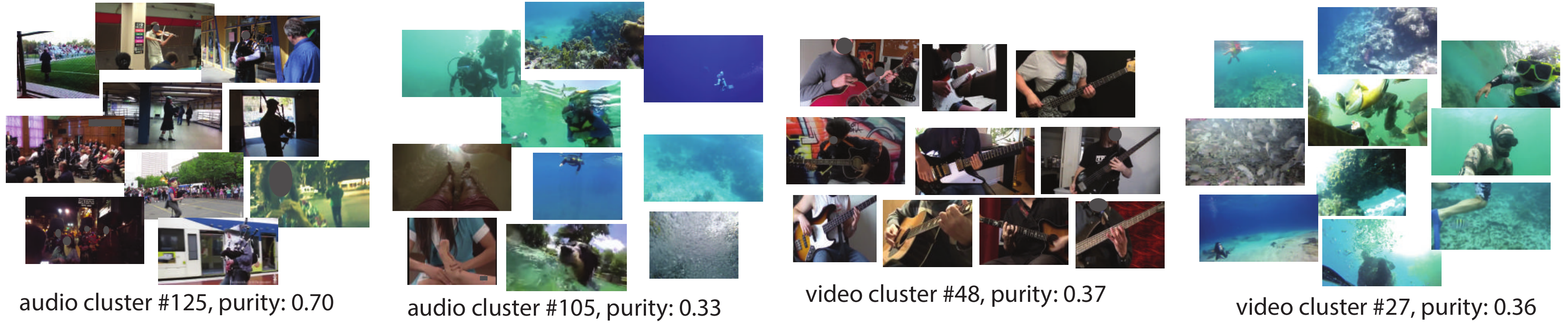}
   \vspace{-16pt}
   \caption{\textbf{Visualization of XDC clusters on Kinetics videos}. The top-$2$ audio clusters (left) and video clusters (right) in terms of purity \emph{w.r.t.} the Kinetics labels. Clusters are represented by the $10$ closest videos (shown as frames) to their centroid. Interestingly, XDC learned to group ``scuba diving'' with ``snorkeling'' (second left, cluster \#105) based on audio features and ``scuba diving'' with ``feeding fish'' (rightmost, cluster \#27) based on visual features. Please refer to our \href{http://humamalwassel.com/publication/xdc/}{project website} for an interactive visualization of all XDC clusters.}
   \label{fig:xdc_clusters}
   \vspace{-10pt}
\end{figure}

\vspace{-3pt}
\section{Understanding XDC}
\vspace{-3pt}
\label{sec:understanding_xdc}

What does XDC actually learn? What semantic signals does the algorithm use to train its encoders? Here, we try to answer these questions by inspecting the $k$-means clustering results produced by the last iteration of XDC. 
Figure~\ref{fig:xdc_clusters} visualizes some audio and video clusters learned by XDC when trained on Kinetics. These clusters are the top $2$ audio clusters (left) and the top $2$ video clusters (right) ranked by purity \emph{w.r.t.} Kinetics action labels. More clusters are presented in Table~\ref{table:xdc_clusters_small}. We observe that the top-purity clusters learned from both modalities exhibit strong semantic coherence. For example, the audio 1st and 8th ranked clusters include concepts related to playing musical instruments that have similar sounds, while the 1st ranked video cluster also groups playing-instrument concepts, but mainly because of their appearance, as the cluster is all about guitars. Other interesting clusters include: grouping by motor-engine sounds (audio \#10), by different swimming strokes (video \#4), by different golf shots (video \#5), and different cooking activities (video \#10). In the bottom-ranked clusters, although the purity \emph{w.r.t.} Kinetics concepts is low, we still find some coherence, mostly at the scene level: a farm setting in audio \#127 (``grooming horse'', ``milking cow'') and gym activities in video \#63 (``pull ups'', ``punching bag''). 
Many other bottom-ranked clusters appear to lack semantic coherence when viewed through the lens of Kinetics labels. However, one of the motivations behind the design of self-supervised methods is precisely to bypass the hand-design of label spaces, which may not be the optimal ones for general representation learning. Our experiments suggest that the label space learned by XDC yields strong and general audio and video features even though it does not align perfectly with the taxonomies of existing datasets. 
\begin{table}[t]
    \small
    \centering
    \tabcolsep=0.05cm
    \caption{{\bf XDC clusters.} Top and bottom audio (left) and video (right) XDC clusters ranked by clustering purity \emph{w.r.t.} Kinetics labels. For each cluster, we list the three concepts with the highest purity (given in parentheses).} 
    \begin{subtable}{.49\linewidth}
        \centering
        \begin{tabular}{ l l}
            \hline
            {\scriptsize \#} & {\scriptsize Kinetics concepts} \\
            \hline
            {\scriptsize 1} & {\scriptsize play bagpipes(0.70), play harmonica(0.04), play violin(0.03)} \\
            {\scriptsize 2} & {\scriptsize scuba diving(0.33), snorkeling(0.27), feeding fish(0.11)} \\
            {\scriptsize 8} & {\scriptsize play cello(0.15), play trombone(0.11), play accordion(0.09)}\\
            {\scriptsize 10} & {\scriptsize mowing lawn(0.14), driving tractor(0.09), motorcycling(0.06)}\\
            \hline
            {\scriptsize 127} & {\scriptsize abseiling(0.01), grooming horse(0.01), milking cow(0.01)}\\
            {\scriptsize 128} & {\scriptsize washing feet(0.01), motorcycling(0.01), headbanging(0.01)}\\
            \hline
        \end{tabular}
    \end{subtable}\hfill
    \begin{subtable}{.49\linewidth}
        \centering
        \begin{tabular}{ l l }
            \hline
            {\scriptsize \#} & {\scriptsize Kinetics concepts} \\
            \hline
            {\scriptsize 1} & {\scriptsize play bass guitar(0.37), play guitar(0.16), tap guitar(0.15)}\\
            {\scriptsize 4} & {\scriptsize swim backstroke(0.21), breast stroke(0.16), butterfly stroke(0.1)}\\
            {\scriptsize 5} & {\scriptsize golf putting(0.18), golf chipping(0.11), golf driving(0.05)}\\
            {\scriptsize 10} & {\scriptsize cook chicken(0.11), barbeque(0.07), fry vegetables(0.06)}\\
            \hline
            {\scriptsize 63} & {\scriptsize pull ups(0.01), gymnastics tumbling(0.01), punching bag(0.01)}\\
            {\scriptsize 64} & {\scriptsize capoeira(0.01), riding elephant(0.01), feeding goats(0.01)}\\ 
            \hline
        \end{tabular}
    \end{subtable}
    \vspace{-8pt}
    \label{table:xdc_clusters_small}
\end{table}

\begin{table}[t]
    \small
	\centering
	\tabcolsep=0.1cm
	\caption{\textbf{State-of-the-art comparison.} We report the average top-1 accuracy over the official splits for all benchmarks. 
	{\bf (a) Video action recognition:} Comparison between XDC with self-supervised and fully-supervised methods on UCF101 and HMDB51. Not only does XDC set new state-of-the-art performance for self-supervised methods, it also outperforms fully-supervised Kinetics and ImageNet pretraining.
	{$^{*}$ For fair comparison with XDC, we report AVTS performance without dense prediction, \emph{i.e.}, we average the predictions of $10$ uniformly-sampled clips at inference.
	$^{\dagger}$ For direct comparison with XDC, we evaluate AVTS using R(2+1)D-18 and $10$ uniformly-sampled clips at inference.}
    {\bf (b) Audio event classification:} We compare XDC with self-supervised methods on ESC50 and DCASE. XDC achieves state-of-the-art performance on DCASE.
	}
    \begin{subtable}{0.66\linewidth}
        \centering
    	\vspace{-4pt}
    	\caption{Video action recognition.}
    	\vspace{-4pt}
    	\begin{tabular}{l l l | c c }
    		\hline
    		\multicolumn{3}{c|}{\underline{Pretraining}} & \multicolumn{2}{c}{\underline{Evaluation}} \\
    		Method & Architecture & Dataset & UCF101 & HMDB51 \\
    		\hline 
    		ClipOrder~\cite{clip_order}                    & R(2+1)D-18 & UCF101 & 72.4 & 30.9 \\
    		\hline
            MotionPred~\cite{motion_statistics} & C3D         & Kinetics & 61.2 & 33.4 \\
    		ST-Puzzle~\cite{kim2019self}        & 3D-ResNet18 & Kinetics & 65.8 & 33.7 \\
    		DPC~\cite{han2019video}             & 3D-ResNet34 & Kinetics & 75.7 & 35.7 \\
            CBT~\cite{sun2019learning}          & S3D         & Kinetics & 79.5 & 44.6 \\
            SpeedNet~\cite{benaim2020speednet}  & S3D         & Kinetics & 81.1 & 48.8 \\
            AVTS~\cite{avts}$^{*}$              & MC3-18      & Kinetics & 84.1 & 52.5 \\
            AVTS~\cite{avts}$^{\dagger}$        & R(2+1)D-18  & Kinetics & 86.2 & 52.3 \\
            {\bf XDC} (ours)                    & R(2+1)D-18  & Kinetics & 86.8 & 52.6 \\
    		\hline
            AVTS~\cite{avts}$^{*}$           & MC3-18     & AudioSet & 87.7 & 57.3 \\
            AVTS~\cite{avts}$^{\dagger}$   & R(2+1)D-18 & AudioSet & 89.1 & 58.1 \\
            {\bf XDC} (ours)            & R(2+1)D-18 & AudioSet & 93.0 & 63.7 \\
            \hline
            MIL-NCE~\cite{miech2020end} & S3D        & HowTo100M & 91.3 & 61.0 \\
            ELo~\cite{piergiovanni2020evolving} & R(2+1)D-50 & YouTube-8M & 93.8 & \underline{67.4} \\
            {\bf XDC} (ours)            & R(2+1)D-18 & IG-Random & \underline{94.6} & 66.5 \\
            {\bf XDC} (ours)            & R(2+1)D-18 & IG-Kinetics & {\bf 95.5} & {\bf 68.9} \\
    		\hline
    		\hline
    		Fully supervised            & R(2+1)D-18 & ImageNet & 84.0 & 48.1 \\
    		Fully supervised            & R(2+1)D-18 & Kinetics & 94.2 & 65.1 \\
    		\hline		
    	\end{tabular}
    	\label{table:sota_video}
    \end{subtable}
    \begin{minipage}{0.32\linewidth}
        \centering
        \vspace{-4pt}
        \begin{subtable}{\linewidth}
            \centering
        	\caption{Audio event classification.}
        	\vspace{-4pt}
        	\label{table:sota_audio}
        	\begin{tabular}{l | c }
        		\hline
        		Method & ESC50  \\
        		\hline
        		Random Forest~\cite{esc50} & 44.3 \\
                Piczak ConvNet~\cite{Piczak2015} & 64.5 \\
                SoundNet~\cite{aytar2016soundnet} & 74.2 \\
                $L^3$-Net~\cite{Arandjelovic17} & 79.3 \\
                AVTS~\cite{avts} & 82.3\\ 
                ConvRBM~\cite{sailor2017} & {\bf 86.5} \\
                \hline
                {\bf XDC} (AudioSet) & 84.8 \\
                {\bf XDC} (IG-Random) & \underline{85.4} \\
        		\hline
        	\end{tabular}
        \end{subtable}\vspace{12pt}
        \begin{subtable}{\linewidth}
        	\centering
        	\begin{tabular}{l | c }
        		\hline
        		Method & DCASE \\
        		\hline 
        		RG~\cite{Rakotomamonjy2015} & 69 \\
        		LTT~\cite{ltt2013} & 72 \\
        		RNH~\cite{rnh2013} & 77 \\
                Ensemble~\cite{stowell2015} & 78 \\
                SoundNet~\cite{aytar2016soundnet} & 88 \\
                $L^3$-Net~\cite{Arandjelovic17} & {93} \\
                AVTS~\cite{avts} & \underline{94} \\
                \hline
                {\bf XDC} (AudioSet) & {\bf 95} \\
                {\bf XDC} (IG-Random) & {\bf 95} \\
        		\hline
        	\end{tabular}
        \end{subtable}
	\end{minipage}
    \vspace{-8pt}
\end{table}

\vspace{-1pt}
\section{State-of-the-art self-supervised learning comparison}
\vspace{-1pt}
\label{sec:compare_sota}

\noindent{\bf Experimental setup.} 
Here, training is similar to our ablations except that we re-train our video encoder on the last clustering assignment using $32$-frame clips. Then following~\cite{avts,Tran18}, we finetune on UCF101 and HMDB51 using $32$-frame clips for both XDC and the fully-supervised baselines. 
Inference is similar to our ablations except for using $32$-frame clips. 
For the audio event classification dataset DCASE~\cite{DCASE}, we follow~\cite{avts} and extract \texttt{conv$\_$5} features for $60$ uniformly-sampled clips per audio sample and learn a linear SVM. We report the average top-1 accuracy over \emph{all splits}. 

\vspace{-1pt}
\noindent{\bf Video action recognition.} Table~\ref{table:sota_video} compares XDC pretrained on four large-scale datasets against state-of-the-art self-supervised methods, after finetuning on the UCF101 and HMDB51 benchmarks\footnote{All XDC pretrained models are publicly released on our \href{http://humamalwassel.com/publication/xdc/}{project website}.}. We also compare against two fully-supervised methods pretrained on ImageNet and Kinetics.
\textbf{\textit{Results:}}
\textbf{{(I)}} XDC pretrained on IG-Kinetics sets new state-of-the-art performance for self-supervised methods on both benchmarks, outperforming Elo~\cite{piergiovanni2020evolving} by $1.7\%$ on UCF101 and $1.5\%$ on HMDB51. Moreover, XDC significantly outperforms fully-supervised pretraining on Kinetics: by $1.3\%$ on UCF101 and by $3.8\%$ on HMDB51.
\textbf{{(II)}} When directly compared on the same R(2+1)D-18 architecture, XDC pretrained on Kinetics slightly outperforms AVTS~\cite{avts} by $0.6\%$ on UCF101 and $0.3\%$ on HMDB51. However, when both methods are pretrained on AudioSet, XDC outperforms AVTS with larger margins: by $3.9\%$ on UCF101 and $5.6\%$ on HMDB51. This shows that XDC scales better than AVTS. To further verify that XDC scales better, we pretrained AVTS on AudioSet-$240$K using R(2+1)D-18 and got $76.9\%$ and $40.7\%$ for UCF101 and HMDB51 on split-1, showing a smaller margin between XDC and AVTS than when both are pretrained on the full AudioSet (cf. Table~\ref{table:ablation_study_pretrain_dataset_type_and_size}).

\vspace{-1pt}
\noindent{\bf Audio event classification.} Table~\ref{table:sota_audio} compares XDC pretrained on AudioSet and IG-Random against the state-of-the-art self-supervised methods for audio classification. XDC achieves state-of-the-art performance on DCASE and competitive results on ESC50 with only a $1.1\%$ gap with~\cite{sailor2017}.

\vspace{-1pt}
\section{XDC for temporal action localization}
\vspace{-1pt}
In this section, we further demonstrate that XDC can be useful beyond video and audio classification. In particular, we employ the recent G-TAD~\cite{xu2020gtad} action localization algorithm, where we replace the clip features (originally extracted from a TSN~\cite{WangXW0LTG16} model pretrained on Kinetics) with our XDC features from the R(2+1)D-18 model pretrained on IG-Kinetics or IG-Random. We compare against the features from the R(2+1)D-18 model fully-supervised pretrained on Kinetics. We emphasize that we do not finetune any of the feature extractors used in this experiment. We follow the default hyperparameters setting of G-TAD. Table~\ref{table:xdc_thumos14} shows temporal action localization results of G-TAD with different features on THUMOS14~\cite{THUMOS14} dataset. It reports the mean Average Precision (mAP) results at different temporal Intersection over Union (tIoU) thresholds. Both XDC variants outperform the fully-supervised features across all tIoU thresholds. This confirms the same trend observed in tasks presented in Section~\ref{sec:compare_sota} and suggests that XDC can be used for other tasks.
\begin{table}[h]
    \small
    \centering
    \vspace{-10pt}
    \caption{{\bf Temporal action localization on THUMOS14.} We compare G-TAD algorithm using our XDC features vs. using the fully-supervised Kinetics-pretrained (Superv) features. We report the mean Average Precision (mAP) results at different temporal Intersection over Union (tIoU) thresholds. Both XDC variants outperform the fully-supervised features across all tIoU thresholds.}
    \begin{tabular}{l | c c c c c }
        \hline
        & \multicolumn{5}{c}{mAP @ tIoU}\\
        Features Type & 0.3 & 0.4 & 0.5 & 0.6 & 0.7 \\
        \hline 
        Superv (Kinetics) & 50.9 & 44.4 & 36.6 & 28.4 & 19.8 \\
        XDC (IG-Random)    & {\bf 51.5} & 44.8 & 36.9 & 28.6 & \textbf{20.0} \\
        XDC (IG-Kinetics) & {\bf 51.5} & {\bf 44.9} & {\bf 37.2} & {\bf 28.7} & {\bf 20.0} \\
        \hline
    \end{tabular}
    \vspace{-10pt}
    \label{table:xdc_thumos14}
\end{table}

\vspace{-6pt}
\section{Conclusion}
\vspace{-7pt}
\label{sec:conclusion}
We presented Cross-Modal Deep Clustering (XDC), a novel self-supervised model for video and audio. XDC outperforms not only existing self-supervised methods but also fully-supervised ImageNet- and Kinetics-pretraining for action recognition. To the best of our knowledge, XDC is the first to show self-supervision outperforming large-scale full-supervision pretraining for action recognition when pretrained on the same architecture and a larger number of uncurated videos.

\section*{Broader Impact Statement}

Video has become a commonplace in society. Its uses range from entertainment, to communication and teaching. Thus, the learning of semantic representations of video has broad and far-reaching potential applications. The authors do not foresee major ethical issues associated to this work.~However, as the proposed approach is self-supervised, it will learn the inherent properties and structure of the training data. Thus, the learned model may exhibit biases intrinsically present in the data.

\section*{Acknowledgments}
The authors thank Mengmeng Xu for his valuable help with the THUMOS14 experiments. The authors appreciate the anonymous NeurIPS reviewers for their constructive feedback. Humam Alwassel was partially supported by the King Abdullah University of Science and Technology (KAUST) Office of Sponsored Research (OSR) under Award No. OSR-CRG2017-3405.

\bibliographystyle{ieee_fullname}
\bibliography{egbib}

\clearpage
\appendix

\section*{Supplementary Material}

\section{Optimization challenges} 
In this section, we give the details of the full optimization cycle and discuss differences between the single-modality baseline and our multi-modal models. 

\noindent {\bf Trivial solutions.}
As discussed in~\cite{caron2018deep}, SDC may converge to trivial solutions, corresponding to empty clusters or encoder parameterizations, where the classifier predicts the same label regardless of the input. DeepCluster proposes workarounds to tackle these issues, involving reassigning empty cluster centers and sampling training images uniformly over the cluster assignments. While these strategies mitigate the issues, they do not fix the main cause of the problem: SDC learns a discriminative classifier on the \textit{same} input from which it learns the labels. On the other hand, our multi-modal deep clustering models are less prone to trivial solutions because they learn the discriminative classifier on one modality and obtain the labels from a \textit{different} modality. In our training, we never encountered the issue of empty clusters or few-class predictions for any of our multi-modal clustering approaches. 

\noindent {\bf Initialization and convergence.} 
Our initial pseudo-labels come from clustering features of randomly-initialized encoders. Such pseudo-labels are ``good enough'' to capture some weak similarities between the input samples as features from randomly-weighted networks have shown decent results on image and audio classification~\cite{pons2019randomly,saxe2011random}. Another potential option involves generating the initial pseudo-labels by clustering hand-crafted features, \emph{e.g.} iDT \cite{Wang2013} and audio spectrograms. Hand-crafted features capture low-level semantics that may help the encoders learn better or faster.
Indeed, in small-scale experiments, we observed that clustering handcrafted features in the initial iteration reduces the number of clustering iterations needed to learn a well-performing encoder. However, we decided to not pursue this further, since these features are computationally expensive to extract and thus are not suitable for large-scale training on millions of examples. Furthermore, handcrafted features may bias the learning to reflect the design choices behind these manually-engineered descriptors. 

\noindent {\bf Clustering and optimization schedule.} 
Following previous work~\cite{caron2018deep}, we cluster the deep features using the $k$-means algorithm primarily for its desirable properties of efficiency and scalability. The number of $k$-means clusters is a key hyperparameter in our framework. Intuitively, using more clusters makes the pretext task harder, as it increases the number of pseudo-classes the classifier must recognize. On the other hand, the diversity of samples to cluster effectively dictates the maximum $k$, for which the grouping is still sensible. Taking into account these factors, we explore the effects of $k$ in our ablation study in Subsection 4.2 of the main manuscript. Another important hyperparameter of our framework is the number of training epochs for the encoders, before re-clustering the learned features. DeepCluster re-clusters after each epoch, which is an expensive design choice when scaling to millions of training samples. Thus, we choose to fix the pseudo-labels and train the encoders until the validation loss for predicting the pseudo-labels saturates. Then, we re-cluster the newly learned features, reassign pseudo-labels, reset the classification layer, and repeat the same process. We find this strategy to be more efficient, as it reduces the number of times we need to invoke $k$-means. 

\section{Learning using audio rather than text from ASR}
We note that while our approach was demonstrated by leveraging audio, the method is general and is easy to adapt to other modalities, including text. While video and text are semantically correlated, audio and video are temporally correlated. Thus, these two form of correlations are likely to provide different forms of self-supervision, potentially leading to further gains when used in combination. A disadvantage of text from ASR is that it is only available for videos with speech. Audio provides information about environmental sounds beyond speech (\textit{e.g.} walking steps, playing guitar, and dog barking) and allows us to train on uncurated datasets of arbitrary Web videos, as we demonstrated with IG-Random.

\section{Hyperparameters and training details}
\noindent {\bf Training}. We train our models using caffe2 with distributed SGD on a GPU cluster, and employ the warmup scheme proposed in~\cite{goyal2017accurate}. The main training parameters are presented in Table~\ref{table:param_def}. We note that the epoch size can be different from the actual number of videos. This is because the total number of clips the model sees during training (with temporal jittering) can be larger than the number of videos.

\begin{table}[t]
    \small
	\centering
	\caption{\textbf{Training parameter definitions.} The abbreviations and descriptions of each training parameters.}
	\begin{tabular}{|l l l|}
		\hline
		Abv. & Name & Description \\
		\hline 
        es & epoch size & The total number of examples the \\
           & & model trains on in one epoch. \\
        \hline
        bs & batch size & The size of a mini-batch. \\
        \hline
        lr & base lr & The initial learning rate. \\
        \hline
        we & warmup epoch & The number of epochs used for \\
           & & warmup~\cite{goyal2017accurate}. \\
        \hline
        se & step epoch & Every se epochs, the learning rate  \\
        $\gamma$ & lr decay & is decayed by multiplying with $\gamma$. \\ 
        \hline
        te & total epoch & The training lasts for te epochs. \\
        \hline
        wd & weight decay & The weight decay used in SGD. \\
        \hline
        e-stop & early stop & Stop training when validation loss \\
               & & is increased in $3$ consecutive epochs. \\
		\hline
	\end{tabular}
	\vspace{-8pt}
	\label{table:param_def}
\end{table}
\begin{table}[t]
    \small
	\centering
	\caption{\textbf{Pretraining parameters.} We use early-stopping for Kinetics and AudioSet since we observe some overfiting on the pretext tasks. For the last iteration of XDC on IG-Kinetics and IG-Random, we pretrain XDC 3x longer (iteration denoted as IG-Kinetics* and IG-Random* in this table). $\gamma$ is set to $0.01$ for all settings.}
	\begin{tabular}{|l | l l l  l l l l |}
		\hline
		method & dataset & es & bs & lr & we/se/te & wd & e-stop \\
		\hline 
    	Superv & Kinetics & 1M & 32 & 0.01 & 10/10/45 & $10^{-4}$ & no \\
    	Superv & AudioSet & 2M & 32 & 0.04 & 10/20/45 & $10^{-5}$ & no \\
    	\hline
        All DCs & Kinetics & 1M & 32 & 0.01 & 10/10/30 & $10^{-4}$ & yes \\
        All DCs & AudioSet & 2M & 32 & 0.01 & 10/10/45 & $10^{-4}$ & yes \\
        All DCs & IG-Kinetics \& IG-Random & 10M & 32 & 0.01 & 1/3/10 & $10^{-4}$ & no \\
        All DCs & IG-Kinetics* \& IG-Random* & 10M & 32 & 0.01 & 0/9/30 & $10^{-4}$ & no \\
		\hline
	\end{tabular}
	\vspace{-8pt}
	\label{table:pretrain_param}
\end{table}
\noindent {\bf Pretraining parameters}. We pretrain XDC and other baselines using the parameters described in Table~\ref{table:pretrain_param}. Early stopping is used for pretraining on small datasets such as Kinetics~\cite{kinetics} and AudioSet~\cite{AudioSet} to stop before the model starts overfitting on the pretext task. For IG-Kinetics~\cite{ghadiyaram2019large} and IG-Random, we do not observe overfitting. We pretrain XDC on IG-Kinetics and IG-Random longer in the last deep clustering iteration (denoted as IG-Kinetics* and IG-Random* in Table~\ref{table:pretrain_param}). When pretraining our R(2+1)D on longer clips (\emph{e.g.} 32 frames), due to the GPU memory limit, we reduce the mini-batch size to $8$ (instead of $32$) and the base learning rate to $0.0025$ (instead of $0.01$).

\begin{table}[t]
    \small
	\centering
	\caption{\textbf{Finetuning parameters.} Different pretraining methods have different ranges of optimal base learning rate when finetuning on downstream tasks. Thus, we cross-validate all methods with the same set of base learning rates and report the best result for each method. $\gamma$ is set to $0.01$ for all settings.}
	\begin{tabular}{|l l l l l l|}
		\hline
		dataset & es & bs & we/se/te & wd & e-stop \\
		\hline 
        HMDB51 & 40K & 32 & 2/2/8 & $0.005$ & no \\
        UCF101 & 106K & 32 & 2/2/8 & $0.005$ & no \\
        ESC50 & 20K & 32 & 2/2/8 & $0.005$ & no \\
		\hline
	\end{tabular}
	\vspace{-8pt}
	\label{table:finetune_param}
\end{table}
\begin{table}[t]
    \small
	\centering
	\caption{\textbf{Finetuning base learning rates.} For a fair comparison, we cross-validate all pretraining methods with the same set of base learning rates. We report the best finetuning result for each method. Learning FC-only benefits from cross-validation with a wider range of base learning rates.}
	\begin{tabular}{|l l|}
		\hline
		Setup & Base learning rates \\
		\hline 
		Full & $0.001, 0.002, 0.004, 0.006, 0.008, 0.01$\\
		FC only & $0.001, 0.002, 0.004, 0.006, 0.008, 0.01, 0.02, 0.04$\\
		\hline
	\end{tabular}
	\vspace{-8pt}
	\label{table:finetune_lr}
\end{table}
\noindent {\bf Finetuning parameters}. We provide finetuning hyperparameters in Table~\ref{table:finetune_param}. Different pretraining methods may have different optimal base learning rate when finetuned on downstream tasks. Thus to make a fair comparison, we cross-validate the finetuning using the same set of base learning rates (presented in Table~\ref{table:finetune_lr}) and report the best result for each pretraining method. As we observed that higher learning rates tend to be beneficial when learning FC-only, we use a wider set of learning rates to cross-validate FC-only models. As done during pretraining, when finetuning R(2+1)D on longer clips (\emph{i.e.} 32 frames), we reduce the mini-batch size to $8$ and reduce the base learning rate to $1/4$ of its original rate.

\section{XDC using a different backbone architecture}
\begin{table}[t!]
    \small
    \centering
    \caption{{\bf XDC using a different backbone.} We present the results of XDC on a different backbone, ResNet3D-18, for the visual encoder. We compare XDC pretrained on Kinetics vs. the two baselines: Scratch and fully-supervised Kinetics-pretraining (Superv) for the same backbone. We report the top-1 accuracy on split-1 of each dataset.}
    \begin{tabular}{l | c c c }
        \hline
        Method & UCF101 & HMDB51 & ESC50 \\
        \hline 
        Scratch (ResNet3D-18) & 60.1 & 25.7 & 54.3 \\
        Superv  (ResNet3D-18) & 87.5 & 54.5 & 82.3 \\
        \hline
        XDC (ResNet3D-18) & 68.0 & 36.3 & 75.5 \\
        \hline
    \end{tabular}
    \label{table:xdc_with_r3d-18}
\end{table}
We pretrain XDC on Kinetics with ResNet3D-18 as the visual backbone and keep the same audio encoder (ResNet-18). The results are compared with those of baselines in Table~\ref{table:xdc_with_r3d-18}. XDC with the ResNet3D-18 backbone outperforms the training from scratch baseline by good margins on three downstream tasks.

\section{Additional qualitative results}
\begin{table}[t]
    \scriptsize
    \centering
    \caption{{\bf XDC audio clusters.} Top and bottom $10$ XDC audio clusters ranked by clustering purity \emph{w.r.t.} Kinetics labels. For each, we list the $5$ concepts with the highest purity (given in parentheses).}
    \begin{tabular}{ l | l }
        \hline
        \# & Kinetics concepts \\
        \hline
        1 & playing bagpipes(0.70), playing 2harmonica(0.04), playing violin(0.03), playing accordion(0.02), marching(0.01)\\
        2 & scuba diving(0.33), snorkeling(0.27), feeding fish(0.11), canoeing or kayaking(0.02), jumping into pool(0.02)\\
        3 & playing cymbals(0.21), playing drums(0.17), marching(0.03), air drumming(0.02), drumming fingers(0.02)\\
        4 & passing American football(0.17), play kickball(0.06), catching or throwing softball(0.05), kick field goal(0.02), sled dog racing(0.02)\\
        5 & presenting weather forecast(0.17), playing poker(0.05), testifying(0.03), tying knot (not on a tie)(0.02), golf putting(0.02)\\
        6 & hurling (sport)(0.17), swimming backstroke(0.05), skiing slalom(0.04), vault(0.03), ski jumping(0.02)\\
        7 & presenting weather forecast(0.15), news anchoring(0.05), filling eyebrows(0.02), braiding hair(0.02), tossing salad(0.02)\\
        8 & playing cello(0.15), playing trombone(0.11), playing accordion(0.09), playing harp(0.07), playing clarinet(0.06)\\
        9 & playing recorder(0.14), playing violin(0.12), playing trumpet(0.08), playing harmonica(0.07), tapping guitar(0.06)\\ 
        10 & mowing lawn(0.14), driving tractor(0.09), motorcycling(0.06), blowing leaves(0.04), water skiing(0.04)\\
        \hline 
        119 & side kick(0.02), front raises(0.01), dunking basketball(0.01), smoking(0.01), high kick(0.01)\\
        120 & clay pottery making(0.02), crawling baby(0.02), brushing teeth(0.01), playing harmonica(0.01), eating spaghetti(0.01)\\
        121 & pushing cart(0.01), hula hooping(0.01), high kick(0.01), blowing out candles(0.01), bench pressing(0.01)\\
        122 & shot put(0.01), feeding birds(0.01), squat(0.01), push up(0.01), high jump(0.01)\\
        123 & opening present(0.01), petting cat(0.01), pushing cart(0.01), washing dishes(0.01), punching bag(0.01)\\
        124 & trimming or shaving beard(0.01), petting cat(0.01), front raises(0.01), massaging back(0.01), tai chi(0.01)\\ 
        125 & feeding birds(0.01), tobogganing(0.01), riding elephant(0.01), feeding goats(0.01), jumping into pool(0.01)\\
        126 & climbing tree(0.01), writing(0.01), archery(0.01), brushing hair(0.01), shining shoes(0.01)\\
        127 & abseiling(0.01), grooming horse(0.01), milking cow(0.01), feeding goats(0.01), juggling balls(0.01)\\
        128 & washing feet(0.01), motorcycling(0.01), headbanging(0.01), cheerleading(0.01), krumping(0.01)\\
        \hline
    \end{tabular}
    \vspace{-8pt}
    \label{table:xdc_full_audio_clusters}
\end{table}

\begin{table}[t]
    \scriptsize
    \centering
    \caption{{\bf XDC video clusters.} Top and bottom $10$ XDC video clusters ranked by clustering purity \emph{w.r.t.} Kinetics labels. For each, we list the $5$ concepts with the highest purity (given in parentheses).}
    \begin{tabular}{ l | l}
        \hline
        \# & Kinetics concepts \\
        \hline
        1 & playing bass guitar(0.37), playing guitar(0.16), tapping guitar(0.15), strumming guitar(0.09), playing ukulele(0.09)\\
        2 & scuba diving(0.36), snorkeling(0.32), feeding fish(0.10), diving cliff(0.02), jumping into pool(0.02)\\
        3 & presenting weather forecast(0.26), playing poker(0.10), news anchoring(0.05), testifying(0.03), giving or receiving award(0.02)\\
        4 & swimming backstroke(0.21), swimming breast stroke(0.16), swimming butterfly stroke(0.10), play ice hockey(0.04), jump into pool(0.04)\\
        5 & golf putting(0.18), golf chipping(0.11), golf driving(0.05), hitting baseball(0.03), archery(0.03)\\
        6 & hurling (sport)(0.17), passing American football (in game)(0.06), skiing slalom(0.04), playing ice hockey(0.03), vault(0.03)\\
        7 & filling eyebrows(0.13), braiding hair(0.05), massaging back(0.05), curling hair(0.05), dying hair(0.03)\\
        8 & playing cello(0.12), playing harp(0.12), playing trombone(0.06), playing piano(0.06), playing accordion(0.05)\\
        9 & windsurfing(0.12), jetskiing(0.10), water skiing(0.09), surfing water(0.08), kitesurfing(0.06)\\
        10 & cooking chicken(0.11), barbequing(0.07), frying vegetables(0.06), cooking sausages(0.04), making pizza(0.04)\\
        \hline
        55 & yoga(0.02), folding napkins(0.02), doing nails(0.02), cutting watermelon(0.01), writing(0.01)\\
        56 & eating spaghetti(0.02), making pizza(0.02), brushing teeth(0.02), blowing out candles(0.02), reading book(0.02)\\
        57 & answering questions(0.02), tai chi(0.02), dancing ballet(0.02), dunking basketball(0.02), sign language interpreting(0.01)\\
        58 & trimming or shaving beard(0.02), barbequing(0.02), kissing(0.02), dining(0.01), playing poker(0.01)\\
        59 & punching bag(0.02), blowing out candles(0.02), pumping fist(0.02), dancing gangnam style(0.02), opening present(0.01)\\
        60 & feeding goats(0.02), blowing out candles(0.02), milking cow(0.02), arm wrestling(0.02), finger snapping(0.02)\\
        61 & air drumming(0.02), pumping fist(0.02), pushing cart(0.02), brushing teeth(0.02), eating ice cream(0.01)\\
        62 & clean and jerk(0.01), robot dancing(0.01), bench pressing(0.01), side kick(0.01), punching bag(0.01)\\
        63 & pull ups(0.01), gymnastics tumbling(0.01), punching bag(0.01), cracking neck(0.01), eating ice cream(0.01)\\
        64 & capoeira(0.01), riding elephant(0.01), feeding goats(0.01), feeding birds(0.01), crawling baby(0.01)\\ 
        \hline
    \end{tabular}
    \vspace{-8pt}
    \label{table:xdc_full_video_clusters}
\end{table}
\noindent{\bf XDC clusters.}
Tables~\ref{table:xdc_full_audio_clusters} and~\ref{table:xdc_full_video_clusters} present the top and bottom 10 audio and video clusters learned with XDC on Kinetics, ranked by their purity with respect to Kinetics labels. We list the $5$ most frequent concepts of each cluster. 

\begin{figure}[t]
  \centering
  \includegraphics[width=\linewidth]{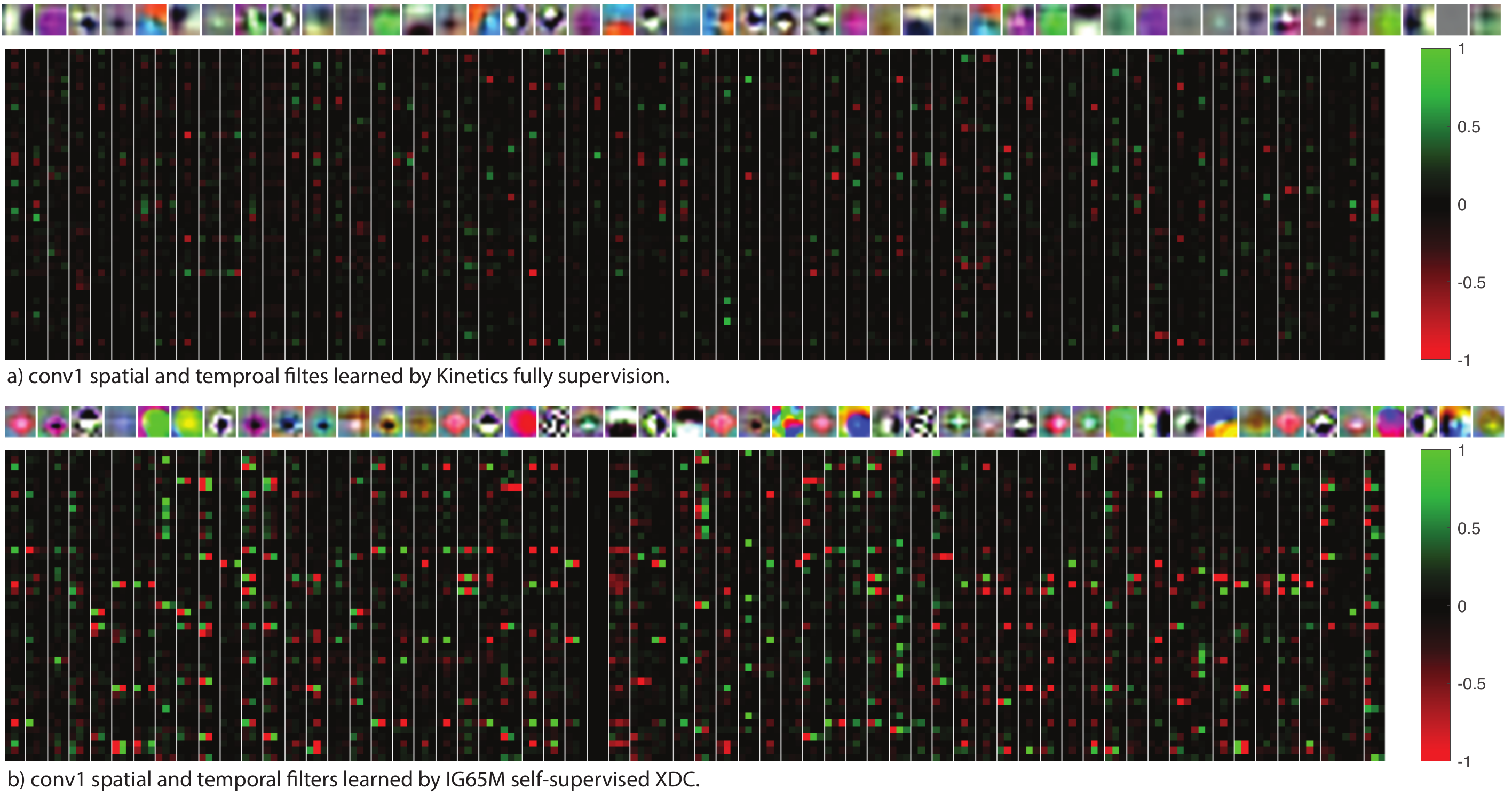}
  \caption{\textbf{R(2+1)D filters learned with self-supervised XDC vs. fully-supervised training.} (a) R(2+1)D \texttt{conv\_1} filters learned by fully-supervised training on Kinetics. (b) The same filters learned by self-supervised XDC pretraining on IG-Kinetics. XDC learns a more diverse set of temporal filters compared to fully-supervised pretraining.
  }
  \label{fig:xdc_filters}
\end{figure}

\noindent{\bf XDC filters.} Figure~\ref{fig:xdc_filters} visualizes and compares \texttt{conv\_1} spatial and temporal filters of R(2+1)D learned by self-supervised XDC pretraining on IG-Kinetics versus fully-supervised pretraining on Kinetics. We observe some differences in both spatial and temporal filters between XDC and fully-supervised pretraining. In particular, XDC learns a more diverse set of motion filters.

\end{document}